\documentclass[conference]{IEEEtran}
\usepackage{graphicx}
\usepackage{booktabs} 			
\usepackage[font=small,skip=0pt]{caption}
\usepackage{lipsum}
\usepackage{stfloats}
\usepackage{algorithm} 
\usepackage{algorithmic} 
\usepackage{multirow} 
\usepackage{amsmath}
\usepackage{caption}
\usepackage{ amssymb }
\usepackage[table]{xcolor}
\graphicspath{ {Figures/} }
\begin{document}

\title{Recent Advances in Features Extraction and Description Algorithms: A Comprehensive Survey}
\author{Ehab Salahat,\textit{ Member, IEEE}, and Murad Qasaimeh, \textit{Member, IEEE}}
\maketitle

\thispagestyle{empty}

\begin{abstract}

Computer vision is one of the most active research fields in information technology today. Giving machines and robots the ability to see and comprehend the surrounding world at the speed of sight creates endless potential applications and opportunities. Feature detection and description algorithms can be indeed considered as the retina of the eyes of such machines and robots. However, these algorithms are typically computationally intensive, which prevents them from achieving the speed of sight real-time performance. In addition, they differ in their capabilities and some may favor and work better given a specific type of input compared to others. As such, it is essential to compactly report their pros and cons as well as their performances and recent advances. This paper is dedicated to provide a comprehensive overview on the state-of-the-art and recent advances in feature detection and description algorithms. Specifically, it starts by overviewing fundamental concepts. It then compares, reports and discusses their performance and capabilities. The Maximally Stable Extremal Regions algorithm and the Scale Invariant Feature Transform algorithms, being two of the best of their type, are selected to report their recent algorithmic derivatives.
\end{abstract}


\section{Introduction} \vspace{-1mm}
Features detection and description from static and dynamic scenes is an active area of research and one of the most studied topics in computer vision literature. The concept of feature detection and description refers to the process of identifying points in an image (interest points) that can be used to describe the image's contents such as Edges, corners, ridges and blobs. It is primarily aiming towards object detection, analysis and tracking from a video stream to describe the semantics of the its actions and behavior \cite{Weiming}. It also has a long list of potential applications, which include, but is not limited to, access control to sensitive building, crowd and population statistical analysis, human detection and tracking, detecting of suspicious actions, traffic analysis, vehicular tracking, and detection of military targets.

In the last few years, we have witnessed a remarkable increase in the amount of homogeneous and inhomogeneous visual inputs (mainly due to the availability of cheap capturing devices such as the built-in cameras in smart phones, in addition to the availability of free image hosting applications, websites and servers such as Instagram and Facebook). This drives the research communities to propose number of novel, robust, and automated features detection and description algorithms, that can adapt to the needs of an application in terms of accuracy and performance.

Most of the proposed algorithms requires intensive computations (especially when it is used with high-definition video stream or with high-resolution satellite imagery applications). Hardware accelerators with massive processing capabilities for these algorithms is required to accelerate the its computations for real-time applications. Digital Signal Processors (DSPs), Field Programmable Gate Arrays (FPGAs), System-on-Chips (SoCs), Application-Specific Integrated Circuits (ASICs), and Graphic Processing Units (GPUs) platforms with smarter, parallelizable, and pipelinable hardware processing designs could be targeted to alleviate this issue.

Porting feature detection and description algorithms into hardware platforms speedup its computation by order of magnitude. However, hardware-constrains such as memory, power, scalability and format interfacing constitute a major bottleneck of scaling it into high resolutions. The typical solution for these hardware-related issues is to scale down the resolution or to sacrifice the accuracy of the detected features. The state-of-the-art in machine and robotic vision, on the other hand, has lately concluded that it is the processing algorithms that will make a substantial contribution to resolve these issues \cite{Liu} \cite{Wang}. That is, computer vision algorithms might be targeted to resolve most of those problems associated with the memory- and power-demanding hardware requirements, and might yield a big revolution for such systems \cite{Ngo}. This challenge is inviting researchers to invent, implement and test these new algorithms, which mainly fall in the feature detection and description category, and which are the fundamental tools of many visual computations applications.

To ensure the robustness of vision algorithms, an essential prerequisite is that they are designed to cover a wide range of possible scenarios with a high-level of repeatability and affine-invariance. Ultimately, studying all of these scenarios and parameters is virtually impossible, however, a clear understanding of all these variables is critical for a successful design. Key factors influencing real-time performance include the processing platform (and its associated constrains on memory, power and frequency in FPGAs, SoCs, GPUs, etc., that can result in algorithmic modifications that can possibly impact the desired performance), monitored environment (e.g. illuminations, reflections, shadows, view orientation, angle, etc.), and the application of interest (e.g. targets of interest, tolerable miss detection/false alarm rates and the desired tradeoffs, and allowed latency). As such, a careful study of computer vision algorithms is essential.

This paper is dedicated to provide a comprehensive overview on the state-of-the-art and recent advances in feature detection and description algorithms. Specifically, the paper starts by overviewing fundamental concepts that constitute the core of feature detection and description algorithms. It then compares, reports and discusses their performance and capabilities. The Maximally Stable Extremal Regions (MSER) algorithm and the Scale Invariant Feature Transform (SIFT) algorithm, being two of the best of their type, are selected to report their recent algorithmic derivatives.

The rest of the paper is organized as follows. Section II provides an overview of the recent state-of-the-art feature detection and description algorithms proposed in literature. It also summaries and compares their performance and accuracy under various transformations. In Section III, the MSER and SIFT algorithms are studied in detail in terms of their recent derivatives. Finally, Section IV concludes the paper with outlooks into future work.

\section{Definitions and Principles}
This section describes the process of detecting and describing a set of feature in an image from a raw image with colored or gray-scale image till the descriptor generation phase. It summarizes the metrics used to measure the quality of the generated feature descriptors.
\subsection{Local features}
Local image features (also known as interest points, key points, and salient features) can be defined as a specific pattern which unique from its immediately close pixels, which is generally associated with one or more of image properties \cite{Tuytelaars} \cite{Li}. Such properties include edges, corners, regions, etc. Figure 1 (a) below represents a summary of such local features. Indeed, these local features represent essential anchor points that can summarize the content of the frame (with the aid of feature descriptors) while searching an image (or a video). These local features are then converted into numerical descriptors, representing unique and compact summarization of these local features.

\begin{figure}[b]  \vspace{-0.5em}
	\centering
	\includegraphics[width=1.0\columnwidth, height=6.3cm]{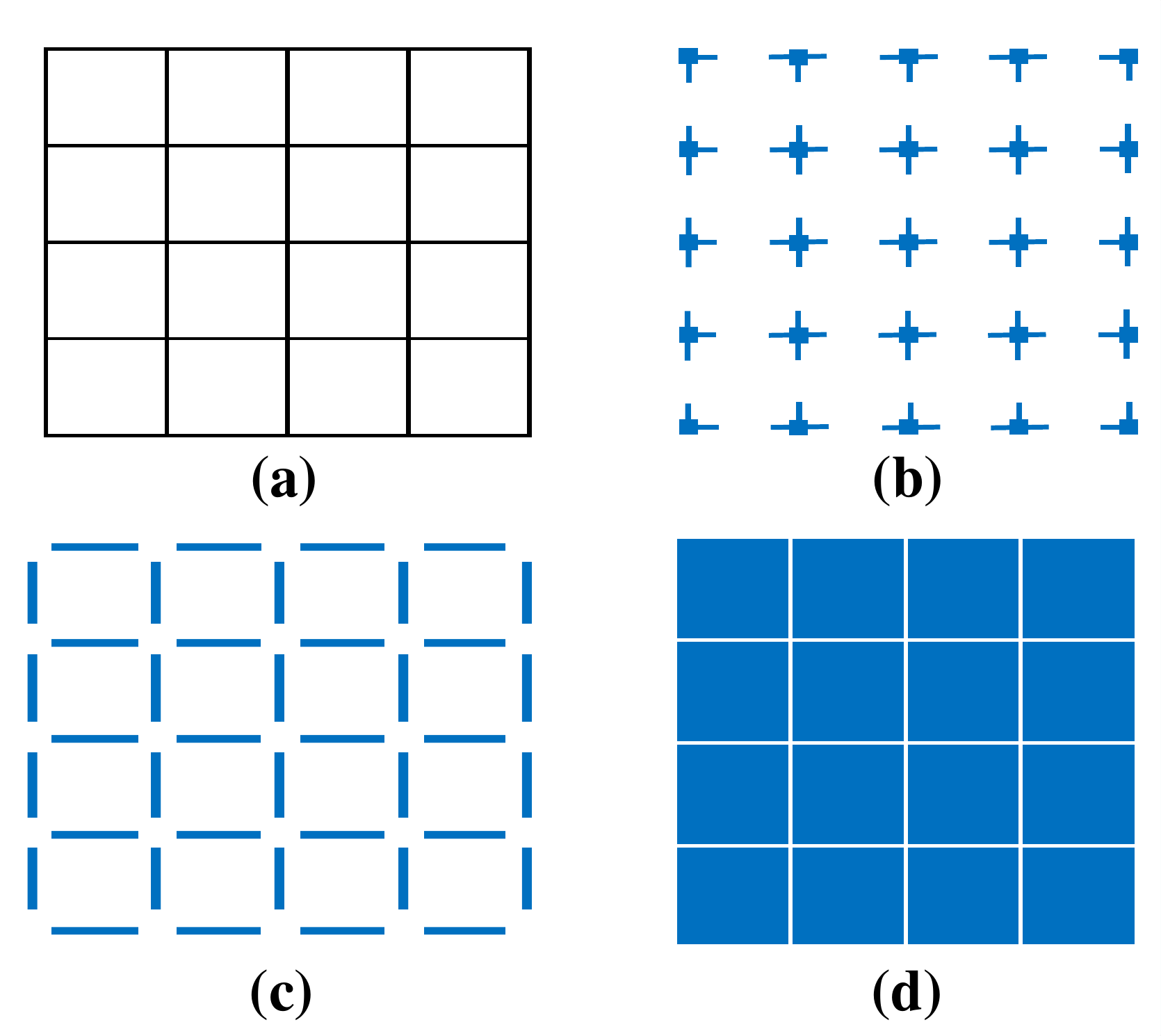}
	\caption{Illustrative image local features (a) input image, (b) corners, (c) edges and (d) regions}
	\label{local_features}
\end{figure}

Local (descriptive and invariant) features provide a powerful tool that can be used in a wide range of computer vision and robotics applications, such as real-time visual surveillance, image retrieval, video mining, object tracking, mosaicking, target detection, and wide baseline matching to name few \cite{Krig}. To illustrate on the usefulness of such local features, consider the following example. Given an aerial image, a detected edge can represent a street, corners may be street junctions, and homogeneous regions can represent cars, roundabouts or buildings (of course, this is a resolution dependent).

The term detector (a.k.a. extractor) traditionally refers to the algorithm or technique that detects (or extracts) these local features and prepare them to be passed to another processing stage that describe their contents, i.e. a feature descriptor algorithm. That is, feature extraction plays the role of an intermediate image processing stage between different computer vision algorithms. In this work, the terms detector and extractor are interchangeably used.

\subsection{Ideal Local Features}

In general, a local feature typically has a spatial extent which is due to its local pixels’ neighborhood. That is, they represent a subset of the frame that is semantically meaningful, e.g. correspond to an object (or a part of an object). Ultimately, it is infeasible to localize all such features as this will require the prerequisite of high-level frame (scene) understanding \cite{Tuytelaars}. As such, those features detection algorithms tries to locate these features directly based on the intensity patterns in the input frame. The selection of these local features can indeed greatly impact the overall system performance \cite{Li}.

Ideal features (and hence feature detectors) should typically have the following important qualities \cite{Tuytelaars}:

\textbf{(1) Distinctiveness}: the intensity patterns underlying the detected features should be rich in variations that can be used for distinguishing features and matching them.

\textbf{(2) Locality}: features should be local so as to reduce the chances of getting occluded as well as to allow simple estimation of geometric and photometric deformations between two frames with different views.

\textbf{(3) Quantity}: the total number of detected features (i.e. features density) should be sufficiently (not excessively) large to reflect the frame’s content in a compact form.

\textbf{(4) Accuracy}: features detected should be located accurately with respect to different scales, shapes and pixels’ locations in a frame.

\textbf{(5) Efficiency}: features should be efficiently identified in a short time that makes them suitable for real-time (i.e. time-critical) applications.

\textbf{(6) Repeatability}: given two frames of the same object (or scene) with different viewing settings, a high percentage of the detected features from the overlapped visible part should be found in both frames. Repeatability is greatly affected by the following two qualities.

\textbf{(7) Invariance}: in scenarios where a large deformation is expected (scale, rotation, etc.), the detector algorithm should model this deformation mathematically as precisely as possible so that it minimizes its effect on the extracted features.

\textbf{(8) Robustness}: in scenarios where a small deformation is expected (noise, blur, discretization effects, compression artifacts, etc.), it is often sufficient to make detection algorithms less sensitive to such deformations (i.e. no drastic decrease in the accuracy).

Intuitively, a given computer vision applications may favor one quality over another \cite{Tuytelaars}. Repeatability, arguably the most important quality, is directly dependent on the other qualities (that is, improving one will equally improve repeatability). Nevertheless, regarding the other qualities, compromises typically need to be made. For example, distinctiveness and locality are competing properties (the more local a feature, the less distinctive it becomes, making feature matching more difficult). Efficiency and quantity are another example of such competing qualities. A highly dense features are likely to improve the object/scene recognition task, but this, however, will negatively impact the computation time.

\begin{table*}[t]
  \centering
  \normalsize
  \caption{A SUMMARY OF THE STATE-OF-THE-ART FEATURE DETECTORS \cite{Li}} \vspace{0.5em}
  \begin{tabular}{| l | l | p{10.5cm} |}  \hline
      \multicolumn{1}{|c|} {Category}   &  \multicolumn{1}{c|} {Classification}	   &   \multicolumn{1}{c|} {Methods and Algorithms} \\  \hline
      Edge-based	 & Differentiation based	& Sobel, Canny      \\  \hline
      Corner-based	& Gradient based	& Harris (and its derivatives), KLT, Shi-Tomasi, LOCOCO, S-LOCOCO \\  \hline
      Corner-based	& Template based	& FAST, AGAST, BRIEF, SUSAN, FAST-ER \\  \hline
      Corner-based	& Contour based   & 	ANDD, DoG-curve, ACJ, Hyperbola fitting, etc. \\  \hline
      Corner-based	& Learning based	& NMX, BEL, Pb, MS-Pb, gPb, SCG, SE, tPb, DSC, Sketch Tokens, etc. \\  \hline
      Blob (interest point)	& PDE based	& SIFT (and its derivatives), SURF (and its derivatives), CenSurE, LoG, DoG, DoH, Hessian (and its derivatives), RLOG, MO-GP, DART, KAZE, A-KAZE, WADE, etc. \\  \hline
      Blob (key point)	& Template based &	ORB, BRISK, FREAK \\  \hline
      Blob (interest region) &	Segmentation based &	MSER (and its derivatives), IBR, Salient Regions, EBR, Beta-Stable, MFD, FLOG, BPLR \\  \hline
  \end{tabular}
  \label{tab:1}
\end{table*}

\subsection{Feature Detectors}
The technical literature is rich with new features detections and description algorithms, and surveys that compare their performance and their qualities such as those mentioned in the earlier section. The reader is referred to some of the elegant surveys from the literature in \cite{Tuytelaars} \cite{Liu8} \cite{Salahat9} \cite{Mikolajczyk10} \cite{Lee11} \cite{Mikolajczyk12} \cite{Miksik13} \cite{Schmid14} \cite{Mikolajczyk15}. However, no ideal detector exists until today. This is mainly due to the virtually infinite number of possible computer vision applications (that may require one or multiple features), the divergence of imaging conditions (changes in scale, viewpoint, illumination and contrast, image quality, compression, etc.) and possible scenes. The computational efficiency of such detectors becomes even more important when considered for real-time applications \cite{Li} \cite{Liu8} \cite{Salahat9}.

As such, the most important local features include: (1) Edges: refer to pixel patterns at which the intensities abruptly change (with a strong gradient magnitude), (2) Corners: refer to the point at which two (or more) edge intersect in the local neighborhood, and (3) Regions: refer to a closed set of connected points with a similar homogeneity criteria, usually the intensity value.

One can intuitively note that there is a strong correlation between these local features. For example, multiple edges sometimes surround a region, i.e. tracking the edges defines the region boundaries. Similarly, the intersection of edges defines the corners \cite{Liu8}. A summary for the well-known feature detectors can be found in table 1. The performance of many of the state-of-the-art detectors is compared in table 2.

As was reported in many performance comparison surveys in the computer vision literature \cite{Tuytelaars} \cite{Mikolajczyk10} \cite{Miksik13}, both the MSER \cite{Matas16} and the SIFT algorithms \cite{Lowe17} have shown an excellent performance in terms of the invariance and other feature qualities (see table 2, the last two rows). Due to these facts, the MSER and SIFT algorithms were extended to several derivatives with different enhancements (that will be reported on later sections). As such, the following section of this paper considers reporting the algorithmic derivatives of the MSER and SIFT algorithms.

\definecolor{blue}{RGB}{217,229,240} \vspace{-0.5em}

\begin{table*}[t]
  \centering
  \normalsize
  \caption{A SUMMARY OF THE PERFORMANCE OF DOMINANT FEATURES DETECTION ALGORITHMS \cite{Tuytelaars}} \vspace{0.5em}
  \begin{tabular}{| l | c | c | c | l | l | l | l |}  \hline
    \multirow{2}{*}{Features Detector} & \multicolumn{3}{c|}{Invariance} & \multicolumn{4}{c|}{Qualities} \\ \cline{2-8}
    & Rotation	& Scale	& Affine	& Repeatability	& Localization	& Robustness	& Efficiency  \\  \hline

    Harris          & $\blacksquare$	& - &	- &	+ + + &	+ + + &	+ + + &	+ +   \\  \hline
    Hessian         &	$\blacksquare$  & -	& - &	+ + &	+ +	& + + &	+             \\  \hline
    SUSAN	          & $\blacksquare$	& - &	- &	+ + &	+ +	& + + &	+ + +         \\  \hline
    Harris-Laplace	& $\blacksquare$	& $\blacksquare$	& -	& + + + &	+ + + &	+ + &	+        \\  \hline
    Hessian-Laplace &	$\blacksquare$	& $\blacksquare$	& - &	+ + + &	+ + + &	+ + + &	+      \\  \hline
    DoG	            & $\blacksquare$	& $\blacksquare$	& - &	+ + &	+ + &	+ + &	+ +          \\  \hline
    Salient Regions	& $\blacksquare$	& $\blacksquare$	& $\blacksquare$	& +	& +	& + + &	+                \\  \hline
    SURF	          & $\blacksquare$	& $\blacksquare$	& -	& + + &	+ + + &	+ +	& + + +      \\  \hline
     \cellcolor{blue}SIFT	          &  \cellcolor{blue}$\blacksquare$  & \cellcolor{blue}$\blacksquare$  & \cellcolor{blue}-	& \cellcolor{blue}+ + & \cellcolor{blue}+ + + & \cellcolor{blue}+ + + & \cellcolor{blue}+ +      \\  \hline
   \cellcolor{blue}MSER	          & \cellcolor{blue}$\blacksquare$  &	\cellcolor{blue}$\blacksquare$  & \cellcolor{blue}$\blacksquare$	& \cellcolor{blue}+ + + &	\cellcolor{blue}+ + + &	\cellcolor{blue}+ +	& \cellcolor{blue}+ + +    \\  \hline

  \end{tabular}
  \label{tab:1}
\end{table*}

\section{MSER AND SIFT: ALGORITHMIC DERIVATIVES} \vspace{0.5mm}

This section discusses some of the well-known MSER and SIFT algorithms derivatives. These algorithms are proposed to enhance the MSER and SIFT algorithms performance in terms of computational complexity, accuracy and execution time.

\subsection{MSER Derivatives}

Maximally stable extremal regions (MSER) algorithm was proposed by Matas et al in 2002. Since then number of region detection algorithms have been proposed based on the MSER technique. The following is a list of five MSER derivatives presented in chronological order.

\textbf{(1) N-Dimensional Extension}: The algorithm was extended first in 2006 for 3D segmentation \cite{Donoser18} by extending the neighborhoods search and stability criteria to 3D image data instead of 2D intensity date. Later on, in 2007, another extension for N-dimensional data space was proposed by Vedaldi in \cite{Vedaldi19}, and later on the same year, an extension to vector-valued function that can be exploited with the three-color channels was also provided in \cite{forssen20}.

\textbf{(2) Linear-Time MSER Algorithm}: In 2008, Nister and Stewenius proposed a new processing flow that emulates real flood-filling in \cite{nister21}. The new linear-time MSER algorithm has several advantages over the standard algorithm such as the better cache locality, linear complexity, etc. An initial hardware design was proposed in \cite{Alyammahi22}.

\textbf{(3) The Extended MSER (X-MSER) Algorithm}: The standard MSER algorithm searches for extremal regions from the input intensity frame only. However, in 2015, the authors of \cite{Salahat23} proposed an extension to the depth (space) domain noting out the correlation between the depth images and intensity images, and introduced the extended MSER detector, which was patented in \cite{Salahat24}.

\textbf{(4) The Parallel MSER Algorithm}: One of the major drawbacks of the MSER algorithm is the need to run it twice on every frame to detect both dark and bright extremal regions. To circumvent on these issues, the authors proposed a parallel MSER algorithm \cite{Salahat25}. Parallel in this context refers to the capability of detecting both extremal regions in a single run. This algorithmic enhancement showed great advantages over the standard MSER algorithm such as a considerable reduction in the execution time, required hardware resources and power, etc. This parallel MSER algorithm has few US patents that are associated with it (e.g. \cite{Salahat26}).

\textbf{(5) Other MSER derivatives}: Other algorithms that were inspired from the MSER algorithm include the Extremal Regions of the Extremal Levels \cite{Faraji27} \cite{Faraji28} algorithm and the Tree-based Morse Regions (TBMR) \cite{Xu29}.

\subsection{SIFT Derievatives}

SIFT algorithm has a local feature detector and local histogram-based descriptor. It detects sets of interest points in an image and for each point it computes a histogram-based descriptor with 128 values. Since SIFT algorithm has been proposed by Lowe in 2004, number of algorithms tried to reduce the SIFT descriptor width to reduce the descriptor computation and matching time. Other algorithms used different window size and histogram compution pattern around each interset point either to speed up the computation process or increase the descripotr robustness against different transformations. One can note that the SIFT is rich with derivatives compared to the MSER algorithm. The reason is that there is not that much to be done to the MSER simple processing flow, unlike the SIFT which is more complicated.  A brief overview of the SIFT algorithmic derivatives are discussed below.

\textbf{(1) ASIFT}: Yu and Morel proposed an affine version of the SIFT algorithm in \cite{Morel30}, which is termed as ASIFT. This derivative simulates all image views obtainable by varying the latitude and the longitude angles. It then uses the standard SIFT method itself. ASIFT is proven to outperform SIFT and to be fully affine invariant \cite{Morel30}. However, the major drawback is the dramatic increase in the computational load. The code of the ASIFT can be found in \cite{ASIFT31}.

\textbf{(2) CSIFT}: Another variation of SIFT algorithm to colored space  is the CSIFT \cite{Abdel32}. It basically modifies the SIFT descriptor (in color invariant space) and is found to be more robust under blur change and affine change and less robust under illumination changes as compared to the standard SIFT.

\textbf{(3) n-SIFT}: The n-SIFT algorithm is simply a straightforward extension of the standard SIFT algorithm to images (or data) with multi-dimensions \cite{Cheung33}. The algorithm creates feature vectors through using hyperspherical coordinates for gradients and multidimensional histograms. The extracted features by n-SIFT can be matched efficiently in 3D and 4D images compared to the traditional SIFT algorithm.

\textbf{(4) PCA-SIFT}: The PCA-SIFT\cite{ke34} adopts an substitute feature vector derived using principal component analysis (PCA), that is based on the normalized gradient patches instead of weighted and smoothed HoG that is used in the standard SIFT. More importantly, it uses a window size 41x41 pixels to generate a descriptor of length 39x39x2= 3042, but  it reduces the dimensionality of the descriptor from 3042 to 20~36 vector by using PCA, which may be more preferable in memory limited devices.

\textbf{(5) SIFT-SIFER Retrofit}: The major difference between SIFT and SIFT with Error Resilience (SIFER) \cite{Mainali35} algorithm is that SIFER (with an improvement in accuracy at the cost of the computational load) has better scale-space management using a higher granularity image pyramid representation and better scale-tuned filtering using a cosine modulated Gaussian (CMG) filter. This algorithm improved the accuracy and robustness of the feature by 20 percent for some criteria. However, the accuracy comes at a cost of increasing the execution time about two times slower than SIFT algorithm.

\textbf{(6) Other derivatives}: Other SIFT derivatives include the SURF \cite{Bay36}, SIFT CS-LBP Retrofit, RootSIFT Retrofit, and CenSurE and STAR algorithms, which are summarized in \cite{Krig}.

\section{Conclusion} \vspace{1mm}

The objective of this paper is to provide a brief introduction for new computer vision researchers about the basic principles of image feature detection and description. It also presents an overview of the recent state-of-the-art algorithms proposed in literature. It starts by reviewing basic yet fundamental concepts that are related to these algorithms. It also provided a brief comparison on their performance and capabilities based on different metrics. The algorithms have been compared in terms of quality of the extracted features under image transformation exist in real-life applications, such as image rotation, scaling and affine. The metrics used in comparisons includes: repeatability, localization, robustness, and efficiency. From this class of algorithms, two of the most frequently used algorithms are selected for detail exploration, MSER and SIFT algorithms with their algorithmic derivatives. The discussion highlighted the derivatives main new aspects that distinguish them from their original form.

\bibliographystyle{ieeetr}
\bibliography{References} 

\begin{thebibliography}{10}

\bibitem{Weiming}
W.~Hu, T.~Tan, L.~Wang, and S.~Maybank, ``A survey on visual surveillance of
  object motion and behaviors,'' {\em IEEE Transactions on Systems, Man, and
  Cybernetics, Part C (Applications and Reviews)}, vol.~34, pp.~334--352, Aug
  2004.

\bibitem{Liu}
H.~Liu, S.~Chen, and N.~Kubota, ``Intelligent video systems and analytics: A
  survey,'' {\em IEEE Transactions on Industrial Informatics}, vol.~9,
  pp.~1222--1233, Aug 2013.

\bibitem{Wang}
G.~Wang, L.~Tao, H.~Di, X.~Ye, and Y.~Shi, ``A scalable distributed
  architecture for intelligent vision system,'' {\em IEEE Transactions on
  Industrial Informatics}, vol.~8, pp.~91--99, Feb 2012.

\bibitem{Ngo}
H.~T. Ngo, R.~W. Ives, R.~N. Rakvic, and R.~P. Broussard, ``Real-time video
  surveillance on an embedded, programmable platform,'' {\em Microprocessors
  and Microsystems}, vol.~37, no.~6, pp.~562--571, 2013.

\bibitem{Tuytelaars}
T.~Tuytelaars, K.~Mikolajczyk, {\em et~al.}, ``Local invariant feature
  detectors: a survey,'' {\em Foundations and trends in computer graphics and
  vision}, vol.~3, no.~3, pp.~177--280, 2008.

\bibitem{Li}
Y.~Li, S.~Wang, Q.~Tian, and X.~Ding, ``A survey of recent advances in visual
  feature detection,'' {\em Neurocomputing}, vol.~149, pp.~736--751, 2015.

\bibitem{Krig}
S.~Krig, ``Computer vision metrics, survey, taxonomy, and analysis. apress,''
  2014.

\bibitem{Liu8}
Q.~Liu, R.~Li, H.~Hu, and D.~Gu, ``Extracting semantic information from visual
  data: A survey,'' {\em Robotics}, vol.~5, no.~1, p.~8, 2016.

\bibitem{Salahat9}
E.~Salahat, H.~Saleh, B.~Mohammad, M.~Al-Qutayri, A.~Sluzek, and M.~Ismail,
  ``Automated real-time video surveillance algorithms for soc implementation: A
  survey,'' in {\em Electronics, Circuits, and Systems (ICECS), 2013 IEEE 20th
  International Conference on}, pp.~82--83, IEEE, 2013.

\bibitem{Mikolajczyk10}
K.~Mikolajczyk and C.~Schmid, ``Scale \& affine invariant interest point
  detectors,'' {\em International journal of computer vision}, vol.~60, no.~1,
  pp.~63--86, 2004.

\bibitem{Lee11}
M.~H. Lee and I.~K. Park, ``Performance evaluation of local descriptors for
  affine invariant region detector,'' in {\em Asian Conference on Computer
  Vision}, pp.~630--643, Springer, 2014.

\bibitem{Mikolajczyk12}
K.~Mikolajczyk and C.~Schmid, ``A performance evaluation of local
  descriptors,'' {\em IEEE transactions on pattern analysis and machine
  intelligence}, vol.~27, no.~10, pp.~1615--1630, 2005.

\bibitem{Miksik13}
O.~Miksik and K.~Mikolajczyk, ``Evaluation of local detectors and descriptors
  for fast feature matching,'' in {\em Pattern Recognition (ICPR), 2012 21st
  International Conference on}, pp.~2681--2684, IEEE, 2012.

\bibitem{Schmid14}
C.~Schmid, R.~Mohr, and C.~Bauckhage, ``Evaluation of interest point
  detectors,'' {\em International Journal of computer vision}, vol.~37, no.~2,
  pp.~151--172, 2000.

\bibitem{Mikolajczyk15}
K.~Mikolajczyk, T.~Tuytelaars, C.~Schmid, A.~Zisserman, J.~Matas,
  F.~Schaffalitzky, T.~Kadir, and L.~Van~Gool, ``A comparison of affine region
  detectors,'' {\em International journal of computer vision}, vol.~65,
  no.~1-2, pp.~43--72, 2005.

\bibitem{Matas16}
J.~Matas, O.~Chum, M.~Urban, and T.~Pajdla, ``Robust wide-baseline stereo from
  maximally stable extremal regions,'' {\em Image and vision computing},
  vol.~22, no.~10, pp.~761--767, 2004.

\bibitem{Lowe17}
D.~G. Lowe, ``Object recognition from local scale-invariant features,'' in {\em
  Computer vision, 1999. The proceedings of the seventh IEEE international
  conference on}, vol.~2, pp.~1150--1157, Ieee, 1999.

\bibitem{Donoser18}
M.~Donoser and H.~Bischof, ``3d segmentation by maximally stable volumes
  (msvs),'' in {\em Pattern Recognition, 2006. ICPR 2006. 18th International
  Conference on}, vol.~1, pp.~63--66, IEEE, 2006.

\bibitem{Vedaldi19}
A.~Vedaldi, ``An implementation of multi-dimensional maximally stable extremal
  regions,'' {\em Feb}, vol.~7, pp.~1--7, 2007.

\bibitem{forssen20}
P.-E. Forss{\'e}n, ``Maximally stable colour regions for recognition and
  matching,'' in {\em Computer Vision and Pattern Recognition, 2007. CVPR'07.
  IEEE Conference on}, pp.~1--8, IEEE, 2007.

\bibitem{nister21}
D.~Nist{\'e}r and H.~Stew{\'e}nius, ``Linear time maximally stable extremal
  regions,'' in {\em European Conference on Computer Vision}, pp.~183--196,
  Springer, 2008.

\bibitem{Alyammahi22}
S.~Alyammahi, E.~Salahat, H.~Saleh, and A.~Sluzek, ``A hardware accelerator for
  real-time extraction of the linear-time mser algorithm,'' in {\em Industrial
  Electronics Society, IECON 2015-41st Annual Conference of the IEEE},
  pp.~65--69, IEEE, 2015.

\bibitem{Salahat23}
E.~Salahat, H.~Saleh, S.~Salahat, A.~Sluzek, M.~Al-Qutayri, and M.~Ismail,
  ``Extended mser detection,'' in {\em IEEE International Symposium on
  Industrial Electronics, \normalfont{Rio de Janeiro, Brazil}}, 3-5 Jun. 2015.

\bibitem{Salahat24}
E.~N. Salahat, H.~H.~M. Saleh, S.~N. Salahat, A.~S. Sluzek, M.~Al-Qutayri,
  B.~Mohammad, and M.~I. Elnaggar, ``Object detection and tracking using depth
  data,'' Oct.~23 2014.
\newblock US Patent App. 14/522,524.

\bibitem{Salahat25}
E.~Salahat, H.~Saleh, A.~Sluzek, M.~Al-Qutayri, B.~Mohammad, and M.~Ismail, ``A
  maximally stable extremal regions system-on-chip for real-time visual
  surveillance,'' in {\em Industrial Electronics Society, IECON 2015-41st
  Annual Conference of the IEEE}, pp.~002812--002815, IEEE, 2015.

\bibitem{Salahat26}
E.~N. Salahat, H.~H.~M. Saleh, A.~S. Sluzek, M.~Al-Qutayri, B.~Mohammad, and
  M.~I. Elnaggar, ``Architecture and method for real-time parallel detection
  and extraction of maximally stable extremal regions (msers),'' Apr.~12 2016.
\newblock US Patent 9,311,555.

\bibitem{Faraji27}
M.~Faraji, J.~Shanbehzadeh, K.~Nasrollahi, and T.~B. Moeslund, ``Erel: extremal
  regions of extremum levels,'' in {\em Image Processing (ICIP), 2015 IEEE
  International Conference on}, pp.~681--685, IEEE, 2015.

\bibitem{Faraji28}
M.~Faraji, J.~Shanbehzadeh, K.~Nasrollahi, and T.~B. Moeslund, ``Extremal
  regions detection guided by maxima of gradient magnitude,'' {\em IEEE
  Transactions on Image Processing}, vol.~24, no.~12, pp.~5401--5415, 2015.

\bibitem{Xu29}
Y.~Xu, P.~Monasse, T.~G{\'e}raud, and L.~Najman, ``Tree-based morse regions: A
  topological approach to local feature detection,'' {\em IEEE Transactions on
  Image Processing}, vol.~23, no.~12, pp.~5612--5625, 2014.

\bibitem{Morel30}
J.-M. Morel and G.~Yu, ``Asift: A new framework for fully affine invariant
  image comparison,'' {\em SIAM Journal on Imaging Sciences}, vol.~2, no.~2,
  pp.~438--469, 2009.

\bibitem{ASIFT31}
J.-M. Morel and G.~Yu, ``Affine-sift (asift).''
  {http://www.cmap.polytechnique.fr/~yu/research/ASIFT/demo.html}, 2016.

\bibitem{Abdel32}
A.~E. Abdel-Hakim and A.~A. Farag, ``Csift: A sift descriptor with color
  invariant characteristics,'' in {\em Computer Vision and Pattern Recognition,
  2006 IEEE Computer Society Conference on}, vol.~2, pp.~1978--1983, IEEE,
  2006.

\bibitem{Cheung33}
W.~Cheung and G.~Hamarneh, ``N-sift: N-dimensional scale invariant feature
  transform for matching medical images,'' in {\em Biomedical Imaging: From
  Nano to Macro, 2007. ISBI 2007. 4th IEEE International Symposium on},
  pp.~720--723, IEEE, 2007.

\bibitem{ke34}
Y.~Ke and R.~Sukthankar, ``Pca-sift: A more distinctive representation for
  local image descriptors,'' in {\em Computer Vision and Pattern Recognition,
  2004. CVPR 2004. Proceedings of the 2004 IEEE Computer Society Conference
  on}, vol.~2, pp.~II--II, IEEE, 2004.

\bibitem{Mainali35}
P.~Mainali, G.~Lafruit, Q.~Yang, B.~Geelen, L.~Van~Gool, and R.~Lauwereins,
  ``Sifer: scale-invariant feature detector with error resilience,'' {\em
  International journal of computer vision}, vol.~104, no.~2, pp.~172--197,
  2013.

\bibitem{Bay36}
H.~Bay, A.~Ess, T.~Tuytelaars, and L.~Van~Gool, ``Speeded-up robust features
  (surf),'' {\em Computer vision and image understanding}, vol.~110, no.~3,
  pp.~346--359, 2008.

\end{thebibliography}

\end{document}